\documentclass[10pt,twocolumn,letterpaper]{article}

\usepackage[pagebackref=true,breaklinks=true,letterpaper=true,colorlinks,bookmarks=false]{hyperref}

\usepackage{cvpr}
\usepackage{times}
\usepackage{epsfig}
\usepackage{graphicx}
\usepackage{amsmath}
\usepackage{amssymb}
\usepackage{breqn}
\usepackage{subfigure} 
\usepackage{adjustbox} 
\usepackage{algorithm,algorithmicx,algpseudocode}

\cvprfinalcopy 

\def\cvprPaperID{7882} 

\title{SoftAdapt: Techniques for Adaptive Loss Weighting of Neural Networks with Multi-Part Loss Functions }

\begin{document}
\author{A. Ali Heydari$^1$\\
{\tt\small aheydari@ucmerced.edu}
\and
Craig A. Thompson$^2$\\
{\tt\small craigthompson@math.arizona.edu}
\and 
Asif Mehmood$^3$\\
{\tt\small asif.mehmood.1@us.af.mil}
}

\maketitle

\begin{abstract}
Adaptive loss function formulation is an active area of research and has gained a great deal of popularity in recent years, following the success of deep learning. However, existing frameworks of adaptive loss functions often suffer from slow convergence and poor choice of weights for the loss components. Traditionally, the elements of a multi-part loss function are weighted equally or their weights are determined through heuristic approaches that yield near-optimal (or sub-optimal) results. To address this problem, we propose a family of methods, called SoftAdapt, that dynamically change function weights for multi-part loss functions based on live performance statistics of the component losses. SoftAdapt is mathematically intuitive, computationally efficient and straightforward to implement. In this paper, we present the mathematical formulation and pseudocode for SoftAdapt, along with results from applying our methods to image reconstruction (Sparse Autoencoders) and synthetic data generation (Introspective Variational Autoencoders).

\end{abstract}

\section{Introduction}
\let\thefootnote\relax\footnote{$^1$ Applied Mathematics Department, University of California, Merced}
\footnote{$^2$ Mathematics Department, University of Arizona}
\footnote{$^3$ Sensors Directorate, U.S. Air Force Research Laboratory}
\footnote{\textbf{Preprint. Under review.}}
Almost all learning through neural networks require (i) a model describing the underlying structure of the training data, (ii) a loss function that gives a metric of how well the network is performing, and (iii) the optimization of the parameters to minimize the objective function. In the past, much of the research had been focused on network architectures \cite{Bengio:2009:LDA:1658423.165842,Koller:2009:PGM:1795555,WeightedGradient}, but recently, more work is being done on how loss functions affect learning \cite{AppleHuang,Barron,Chen}. Networks that perform challenging tasks or multiple tasks often require a combination of losses. Multiple losses are typically combined by taking an equally-weighted linear combination of each objective function; but the importance of each part could be different and thus components should be assigned weights as per their contribution to the learning. On the other hand, the scaling of each component of the loss function can inhibit the ability of the optimizer by only looking at loss components with the largest magnitude. The scaling for gradient descent-based optimizers has been a known issue \cite{Jamil2013ALS}, which our algorithm tries to address throughout the training.

 \begin{figure}[]
\centering

\subfigure[Fixed Weights] {\includegraphics[width=0.14\textwidth]{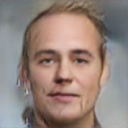}}%
\hfill
\subfigure[SoftAdapt] {\includegraphics[width=0.14\textwidth]{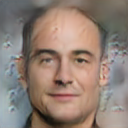}}%
\hfill
\subfigure[Target] {\includegraphics[width=0.14\textwidth]{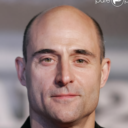}}\\

\subfigure[Fixed Weights] {\includegraphics[width=0.14\textwidth]{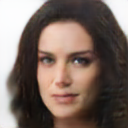}}%
\hfill 
\subfigure[SoftAdapt] {\includegraphics[width=0.14\textwidth]{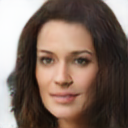}}%
\hfill
\subfigure[Target]{\includegraphics[width=0.14\textwidth]{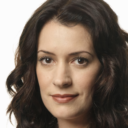}}%

\caption{Reconstruction of the target image after 250 epochs using IntroVAE by Huang et al. \cite{Huang2018}. \textbf{(a), (d)}:  fixed ``optimal" loss function weights ($\alpha,\beta$) that Huang et al. found.  \textbf{(b), (e)}: SoftAdapt adaptive weight balancing. SoftAdapt outperforms the ``optimal" weights in different metrics, described in Section 4.2}

\end{figure}

 In recent years, the need for weighting the components of multi-part functions has become more evident, and researchers have tried to develop different methods to adjust the weights on the linear combination of loss components. These methods often require defining new loss functions \cite{Barron} or changing the optimization procedure \cite{Chen}, but there is limited research on the formulation of a general method that can be added to existing architectures. In most cases, the integration with the current models requires sophisticated adjustment or much longer computation time. The advantage of our method is  compatibility with any gradient descent-based optimizers in machine learning. Our algorithm can also be used in other optimization applications; for example, in convex optimization, the inverse of the Hessian is a popular preconditioner for gradient descent \cite{Li_2018}. However, the Hessian may not be readily available for different applications (\eg in machine learning). Our method may be viewed as using the previous/initial iterations to create a preconditioner matrix $P$ that is a diagonal matrix, such that $P\nabla g$ is approximately isotropic in the parameter space, where $\nabla g$ is a partial gradients of the objective function.

In autoencoders (AE), where the goal is to reconstruct the input data using an encoder and a decoder, a regularization term can be added to the default reconstruction loss. This would encourage the model to have different properties (such as sparsity of the representation or robustness to noise). On the other hand, in variational autoencoders (VAE) \cite{Kingma2013AutoEncodingVB}, where the goal is to generate new data that is similar to the input, the two loss functions are Mean Squared Error ($MSE$) and Kullback-Leibler ($\mathcal{KL}$) divergence (assuming that the prior distribution is a Gaussian). In the case of VAEs, the two losses are crucial for the reconstruction of the input data and estimating the prior distribution to generate new samples; but in AEs the regularization may play a different role in training depending on the problem. An equally weighted linear combination of the losses would mean that each part of the loss function is equally as important in training, which is often not the case \cite{Chen}. For example, sparse autoencoders \cite{Makhzani2013} employ a very small fixed weight ($\ll 1$) on the regularization term as the sparsity parameter, often denoted by $\lambda$, which is usually found by trial and error. Our methods provide learnable parameters that are not fixed, \ie they adapt depending on the performance of the model.

In this paper, we propose a set of Softmax-inspired methods that will adaptively update the weights of the linear combination of individual objective functions, depending on the performance of each part and the collective loss function as a whole.  Our family of techniques, called SoftAdapt, can be thought of as ``add-ons": one can use their choice of optimizer and only add the weights to the linear combination of the losses, as long as both the losses and the optimizers are suitable for the problem at hand. SoftAdapt evaluates the performance by approximating the rate of change of each loss function over a short history, which indicates if it has been increasing or decreasing. SoftAdapt then compares the individual rates of change and determines how visible each objective function should be to the optimizer.

In summary, our contribution is a family of methods that dynamically learn the best weighting on each part of a multi-component loss function, based on live performance metrics. SoftAdapt is fast, easy to implement and can be added to existing architectures that use any gradient descent-based optimizer.

\section{Related Work} \label{sec:2}
Multi-task learning, where the model tries to minimize multiple objective functions to produce an output, is necessary for more challenging tasks but are hard and expensive to train. This learning regime has a wide range of applications, from traffic prediction (\cite{TrafficHuang}) to natural language processing (\cite{Collobert,PlankSG16}), and it was introduced even before the exploration of deep learning (\cite{Bakker:2003:TCG:945365.945370,Caruana1998}). After the deep learning surge, most researchers studied various architectures for multi-objective networks, but more recently, some work has been done towards improving the optimizing of multi-tasking network, based on the optimization functions. Chen et al. \cite{Chen} contemplate on normalizing the gradients for classification and regression tasks in computer vision.

Miranda and Von Zuben \cite{WeightedGradient} explored the problems and limitations of equally-weighted linear combinations for multi-objective loss functions. Similar to our approach, they interpreted machine learning from a multi-objective optimization perspective. However, they introduced an alternative way of optimization using the gradient of the hypervolume, which is defined as the weighted mean of individual loss gradient. Our family of methods calculate the weights on the linear combination adaptively using the exact gradients computed by any traditional optimizer.

In 2019, Xu et al. \cite{Xu2018AutoLossLD} studied the importance of component-wise weighting of the loss function; they designed AutoLoss, a framework that learns and determines the scheduling of the optimization. Very similar to our approach, they realized that in multi-task learning it is important to dynamically set a schedule of training, depending on the network architecture. Our techniques use a different metric to find the importance of optimizing each element of the loss function and can be applied to any multi-part loss function. SoftAdapt can also be interpreted as a scheduling algorithm, but it does not assign discrete weights to the component losses. Our algorithm is fast and generalizable to any multi-part loss function since it uses an approximation to the rate of change of each part using a short history, and it is agnostic to the method of training or type of architecture (\eg Autoencoders, GANs, \etc)  

\section{Methods and Approach}
\label{sec:3}
In this section, we first discuss the mathematical intuition behind multi-part loss weighting and the basic ingredients required for its formulation. Then we will discuss our algorithm SoftAdapt with its two normalized forms, and we provide pseudocode for the implementation. 

\subsection{Mathematical Formulation}
Consider a loss function of the form
\begin{equation}
F(x) = \sum_{k=1}^{n} f_k(x) \text{for} 
\end{equation}
where we wish to minimize $ F $ w.r.t. $  {x} \in \mathbb{R}^m $. Let us suppose that we wish to utilize with some gradient based optimizer $  {x}^{i+1} = Q( {x}^i,g^i) $, where $ g^i $ is the stepping direction for $  {x}^i $. Typically, we take $ g^i = \nabla F( {x}^i) $. In general, without computing additional information, one could have $ g^i $ be dependent on past values of $  {x}^i $ as well as the component losses ($ f_k( {x}^j)$) and the gradients of the component losses ($ \nabla f_k( {x}^j) $), where iteration are denoted by $ j=0,\dots,i $ and component by $ k=1,\dots,n $. There are several methods which take advantage of gradient and step information from previous time-steps (\eg Momentum \cite{Momentum-Backprop}, AdaGrad \cite{AdaGrad}, Adam \cite{Kingma2014AdamAM}, etc), but few, if any, consider recombining the component loss functions; SoftAdapt is designed to address this issue. Let our modified step direction $ h^i $ to be given by
\begin{equation}
h^i = \sum_{k=1}^n \alpha^i_k \nabla f_k( {x}^i)
\end{equation}
and substitute this into $ Q $ in place of $ g^i $. We compute the weights $ \alpha_k^i $ according to previous loss information. There are three main variations for computing $\alpha_k^i$.
\subsubsection{Original Variant (SoftAdapt)}
Here we use the heuristic that it is better to favor the gradient of a function according to its recent performance. Let $ s_k^i $ be an approximation of the recent rate of change of the component loss $ f_k^i:= f_k( {x}^i) $ (\eg $ s_k^i = f_k^i-f_k^{i-1} $, or a more accurate finite difference approximation). Then take
\begin{equation}\label{eq-original_sa}
\alpha_k^i = \frac{e^{\beta s^i_k}}{\sum_{\ell=1}^ne^{\beta s^i_\ell}},
\end{equation}
where $ \beta $ is a tunable hyper-parameter. If one chooses $ \beta > 0 $, SoftAdapt will assign more weight to the worst performing component of the loss function (\ie the component with most positive rate of change). Setting $ \beta < 0 $ favors the best performing losses (most negative rate of change). Taking $ \beta=0 $ gives equal weights. This is simply the classic Softmax evaluation of the vector $ (s_1^i,\dots,s_n^i) $, and is where the method, SoftAdapt, gets its name.

\subsubsection{Loss Weighted}
Here we modify the Softmax function to account for the current values of the losses, as well as their rates of change. Let
\begin{equation}\label{eq:loss_weighted_sa}
\alpha_k^i = \frac{f_k^ie^{\beta s^i_k}}{\sum_{\ell=1}^nf_\ell^i e^{\beta s^i_\ell}}.
\end{equation}
For loss weighting, the component losses must share a minimum (in general, have intersecting minimal sets). The advantage of using this variant is in assigning smaller weights to functions that are close to their minima, even if rates of change stay constant or positive.

\subsubsection{Normalized}
If one wishes, they may normalize the vector $ (s_1^i,\dots,s_n^i) $ before applying it in Eq. \eqref{eq-original_sa} or Eq. \eqref{eq:loss_weighted_sa}. This has the effect of sharpening the distinction between small rates of change and softening it between large ones. Normalized and Loss Weighted may be used together if much smaller weights are desirable for loss functions near their minima.

\subsection{SoftAdapt}
\begin{algorithm}[H]
          \caption{Pseudocode for a SoftAdapt and variations: This algorithm is based on loss function $L$ to be comprised of multiple losses. In general, let $ L = l_1 + l_2 + \cdots + l_m$ }
          \begin{algorithmic}[1] 
          
            \Require{$n$: number of loss values to be stored}
            \Require{Optimizer: An optimizer for the gradient descent-based method}
            \Require{$l_i$: the values of the individual $m$ loss functions}
            \Require{variant: A list of variants to be applied to SoftAdapt. A potentially empty subset of \{``Normalized", ``Loss Weighted"\}}
            \Require{$\epsilon = 10^{-8}$ for numerical stability  }

            \Ensure{$n$ many epochs/iterations have passed before calling \textit{SoftAdapt}}
            \Ensure{$n$ many $l_i$ have been stored for each $l_i$}

            \While{not converged}
                \State{$\beta \gets 0.1$ (default value that can be changed)}\smallskip
				 \State{\(s_i \gets\) the rate of change (up to $(n-1)$th order accurate) of the past $l_i$} \smallskip
				 \State{$f_i \gets$ the average of up to $n$ previous $l_i$} \smallskip

				  				  
				 
              \If{ variant contains ``Normalized" }\smallskip
                \State{$\displaystyle ns_i \gets  \frac{s_i}{\left(\sum_{i=1}^m |s_i| \right) + \epsilon} $} \smallskip
                \State{$\displaystyle \alpha_i = \frac{e^{\beta(ns_i - \max(ns_i))}}{\left(\sum_{j = 1}^m e^{\beta(ns_j - \max(ns_j))} \right) + \epsilon}$}\smallskip

              \Else\smallskip
                \State{$\displaystyle \alpha_i = \frac{e^{\beta(s_i - \max(s_i))}}{\left(\sum_{j = 1}^m e^{\beta(s_j - \max(s_j))}\right) + \epsilon}$}\smallskip
              \EndIf

              \If{ variant contains ``Loss Weighted" }\smallskip
                \State{$\displaystyle \alpha_i = \frac{ f_i \alpha_i}{\left(\sum_{j = 1}^m f_j \alpha_j \right) + \epsilon}$}

              \EndIf
              	\smallskip
          	  \State{\(TLoss \gets l_1 + l_2 + \cdots l_m ~\) (true loss for performance measurer)}\smallskip
			 \State{\(WLoss \gets \alpha_1 l_1 + \alpha_2 l_2 + \cdots \alpha_m l_m ~\)(weighted loss for the optimizer)} 
			\State{optimizer (\(WLoss\))} 

             \EndWhile

          \end{algorithmic}
        \end{algorithm}

\section{Experiments and Results}
\label{sec:3}
In this section, we conduct various experiments to evaluate the performance of SoftAdapt in different test cases. First, we test Original and Loss Weighted SoftAdapt on the \textit{Rosenbrock} function \cite{Rosenbrock} and Beale's function \cite{Jamil2013ALS} (in \textit{Supplementary Material} section) using gradient descent. Then, we will test our proposed method on an Introspective Variational Autoencoder (IntroVAE) \cite{Huang2018} that uses fixed weights. Lastly, we examine SoftAdapt on a Sparse Autoencoder (SAE) \cite{Makhzani2013} to find the sparsity parameter dynamically during training. For both IntroVAE and SAE, we only change the weighting on the loss components using Loss Weighted SoftAdapt while optimizing with Adam \cite{Kingma2014AdamAM}.

\subsection{Gradient Descent Optimization}

As an initial experiment, the SoftAdapt algorithm was tested on a simple gradient descent of standard functions of real vectors. Formally, the minimization problem is: given a smooth function $f:\mathbb{R}^n\to\mathbb{R}$, find an input $ {x}$ which minimizes it, or
\begin{equation}
 {x}=\arg\min_{ {y}\in\mathbb{R}^n}(f( {y}))
\end{equation}

In this section, we present our results on applying the SoftAdapt modified gradient to the classical gradient descent algorithm, both with fixed step size and adaptive step size. The first function to consider is the 2D Rosenbrock function \cite{Rosenbrock}:
\begin{equation}
f(x,y) = (1-x)^2 + 100(y-x^2)^2
\end{equation}
which exhibits a narrow valley that leads to a single global minimum at $ (x,y) = (1,1) $. Typically, gradient descent on the Rosenbrock function will either diverge at step sizes on the order of $10^{-2}$ and larger, or will take a long time to converge. For the experiment, we split $f(x,y) = f_1(x,y) + f_2(x,y)$ where
\begin{equation}
f_1(x,y) = (1-x)^2 \quad \text{and} \quad f_2(x,y) = 100(y-x^2)^2
\end{equation}
For our update procedure we consider two cases. First, we have normal gradient descent:
\begin{equation}
 {x}^{i+1} =  {x}^i - \eta h^i
\end{equation}
where $\eta$ is the fixed learning rate, and $h^i$ is one of the SoftAdapt variant gradients. Second, we will use an adaptive learning rate:
\begin{equation}  {x}^{i+1} =  {x}^i - \eta^i h^i \end{equation}
where $\eta^i$ is updated according to the Barzilai-Borwein scheme \cite{barzilai1988two}, subject to a minimum and maximum learning rate.
\begin{figure*}[!ht]
    \centering
    \subfigure[SoftAdapt, fixed $lr$]{
    \includegraphics[width=0.36\linewidth]{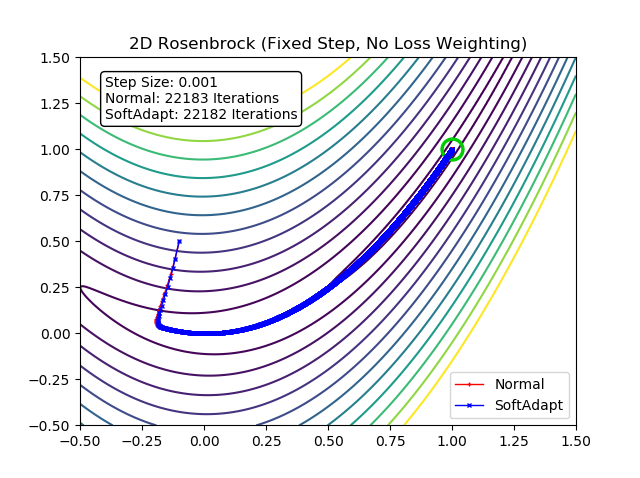}}
    \subfigure[Loss Weighted SoftAdapt, fixed $lr$]{
    \includegraphics[width=0.36\linewidth]{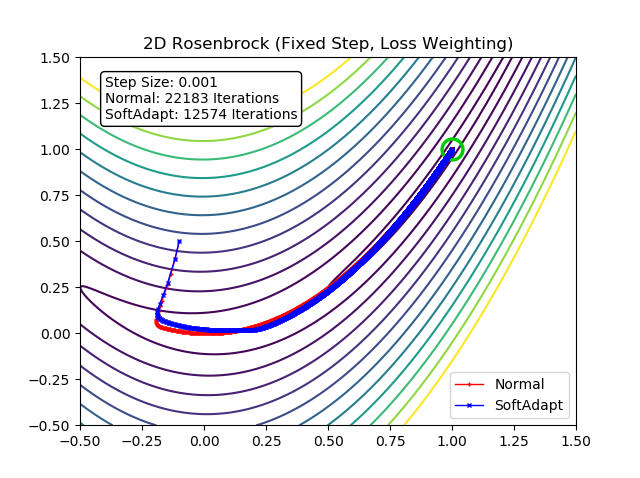}} \\
    \subfigure[SoftAdapt, adaptive $lr$]{
    \includegraphics[width=0.36\linewidth]{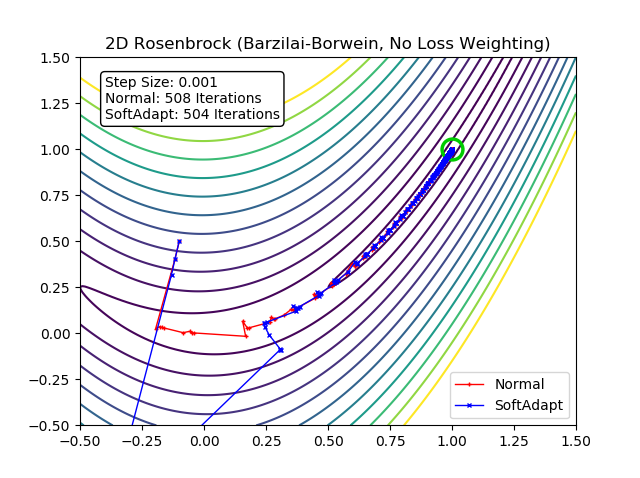}}
    \subfigure[Loss Weighted SoftAdapt, adaptive $lr$]{
    \includegraphics[width=0.36\linewidth]{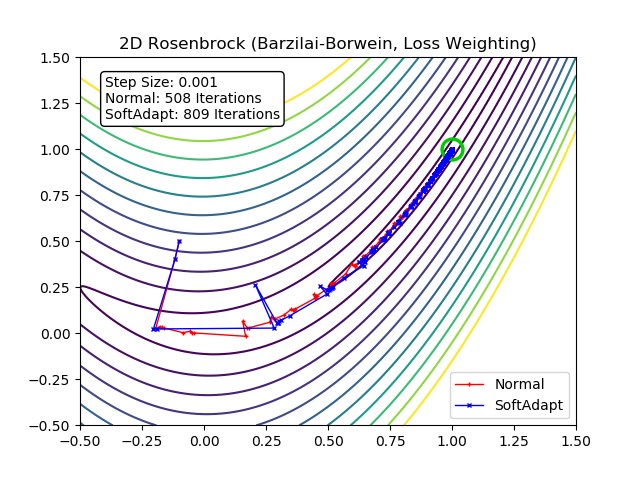}}
    \caption{ Performance of SoftAdapt vs. gradient descent for the Rosenbrock function. The learning rate ($lr$) is changing according to the Barzilai-Borwein scheme \cite{barzilai1988two} in \textbf{(b), (c)}. We see the most improvement ($43.31\%$ faster) for loss weighted SoftAdapt with fixed learning rate. Upon changing the value of $\beta$, significant improvements can be made, but the default values of parameters in our implementation are $\beta = 0.1$, $\eta = 10^{-3} $. The max and min $lr$ are $\eta_{min} = 10^{-4}$, $\eta_{max} = 10^{-1}$}
    \label{fig:grad-vs-softadapt}
\end{figure*}
Fig. \ref{fig:grad-vs-softadapt} shows the trajectories for both, traditional gradient descent and gradient descent with SoftAdapt, and the number of steps taken to reach the minimum. Note that it is appropriate to use loss weighting here, as the minimal sets of $f_1$ and $f_2$ intersect at the true minimum. Our method performs well in three of the regimes and significantly outperforms gradient in fixed step, loss weighting (Fig. \ref{fig:grad-vs-softadapt} \textbf{(b)}). We underperform in the case where both adaptive learning rates and loss weighting are used, but using values of $\beta < 0$ can improve performance. We also witnessed similar improvements in the gradient descent optimization for Beale's function, which is illustrated in \textit{Supplementary Material}.
\subsection{Introspective Variational Autoencoders }

Introspective Variational Autoencoder (IntroVAE) was first introduced by Huang et al. \cite{Huang2018} in 2018. IntroVAE is a single-stream generative model that self-evaluates the quality of the generated images, as opposed to Generative Adversarial Networks\cite{Goodfellow:2014:GAN:2969033.2969125} (GAN), which have a separate network for generating samples and a separate network for discriminating between real and synthetic images. Their interesting approach is that ``[IntroVAE] inference and generator models are jointly trained in an introspective way. On one hand, the generator is required to reconstruct the input images from the noisy outputs of the inference model as normal VAEs. On the other hand, the inference model is encouraged to classify between the generated and real samples while the generator tries to fool it as GANs." \cite{Huang2018}. In the model, the authors use the following loss functions for the encoder (denoted by $L_E$) and for the generator (denoted by $L_G$):  

\begin{equation}\label{eq:OG-LE}
    L_E = L_{REG}(z) + \alpha \sum_{s=r,p} [m-L_{REG}(z_s)]^{+} + \beta L_{AE}(x,x_r)
\end{equation}  

\begin{equation}\label{OG-LG}
L_G = \alpha\sum_{s=r,p} L_{REG}(Enc(x_s)) + \beta L_{AE}(x,x_r) 
\end{equation}  
where $L_{REG}$ is the $\mathcal{KL}$-divergence, which can be computed for $N$ data samples (with dimension of $z$ as $M_z$) as : \\
\begin{equation}
L_{REG}(z;\mu,\sigma) = \frac{1}{2} \sum_{i=1}^N \sum_{j=1}^{M_z} (1+\log(\sigma_{i,j}^2) - \mu_{i,j} - \sigma_{i,j}^2)
\end{equation} 
$L_{AE}$ is the mean squared error: given $x_r$ (the reconstructed image of $x$) and the dimension of $x$ as $M_x$, we have : 

\begin{equation}\label{OG-LAE}
L_{AE}(x,x_r) = \frac{1}{2}\sum_{i=1}^N \sum_{j=1}^{M_x} \|x_{r,ij} - x_{ij}\|_{F}^2 
\end{equation} 
In Eq. (\ref{eq:OG-LE}), $m$ is a number which is selected to keep $L_{REG} $ below a threshold and $Enc(\cdot)$ represents function that the encoder is mapping. For this paper, our focus is on the $\alpha$ and $\beta$, which the authors note as the ``weighting parameters used to balance the importance of each item." \cite{Huang2018}

Our results show that the optimal set of $\alpha$ and $\beta$ does not need to be known in advance since the importance of each part of the loss function can be determined adaptively throughout training using our method. Huang et al. find the ``optimal" value of $\alpha$ and $\beta$ empirically and by pre-training the networks for each different dataset; this results in a different set of $\alpha$ and $\beta$ for different data. The authors make note of this issue and provide the readers with a set of values for each subset of the CELEBA dataset \cite{liu2015faceattributes}. One can avoid finding these weights explicitly for various training data by using SoftAdapt instead since the weight would be learned adaptively during training. Using SoftAdapt, the weighted loss functions in Eq. (\ref{eq:OG-LE}), (\ref{OG-LAE}) will be 
\begin{dmath}
L_E^{(n+1)} = L_{REG}(z) + \alpha_1^{(n)} \sum_{s=r,p} [m - L_{REG}(z_s)]^{+} \newline
+ \alpha_2^{(n)} L_{AE}(x,x_r)
\end{dmath}

\begin{equation}\label{NEW_LE}
L_G^{(n+1)} = \alpha_1^{(n)} \sum_{s=r,p} L_{REG}(Enc(x_s))  + \alpha_2^{(n)} L_{AE}(x,x_r)
\end{equation}

using \begin{equation}\label{alphas}
\alpha_{i}^{(n)} = SoftAdapt\left(L_{REG}^{(n)},L_{AE}^{(n)}\right)
\end{equation}

where $i=\{1,2\}$ and $n \in \mathbb{N}$ denoting the time step for $\alpha_i$. We initialize $\alpha_i^{(0)}=0.5$ since we want to treat it without bias in the very beginning.\\

\begin{figure*}[!ht]
\centering

\subfigure[Epoch 12]
{\includegraphics[width=0.13\textwidth]{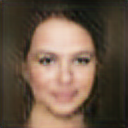}}%
\hspace{9pt} 
\subfigure[Epoch 100]
{\includegraphics[width=0.13\textwidth]{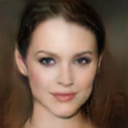}}%
\hspace{9pt}
\subfigure[Epoch 200]
{\includegraphics[width=0.13\textwidth]{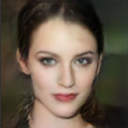}}%
\hspace{9pt}
\subfigure[Epoch 250]
{\includegraphics[width=0.13\textwidth]{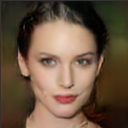}}%
\hspace{30pt}
\subfigure[Target]{\includegraphics[width=0.13\textwidth]{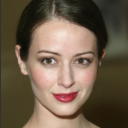}}\\

\subfigure[Epoch 12] {\includegraphics[width=0.13\textwidth]{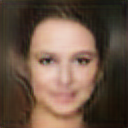}}%
\hspace{9pt}
\subfigure[Epoch 100] {\includegraphics[width=0.13\textwidth]{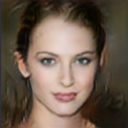}}%
\hspace{9pt}
\subfigure[Epoch 200] {\includegraphics[width=0.13\textwidth]{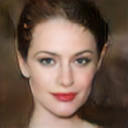}}%
\hspace{9pt}
\subfigure[Epoch 250] {\includegraphics[width=0.13\textwidth]{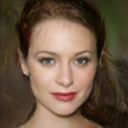}}
\hspace{30pt}
\subfigure[Target]{\includegraphics[width=0.13\textwidth]{female1_target.png}}
\caption{Reconstruction of the target image [\textbf{(e)}, \textbf{(j)}] using IntroVAE with fixed loss weighting from Huang et al. \cite{Huang2018} [images \textbf{(a-d)}] vs our adaptive loss weighting with SoftAdapt [images \textbf{(f-i)}]} 

\label{fig:female1SA}
\end{figure*}

\begin{figure*}[!ht]
\centering

\subfigure[Epoch 12]
{\includegraphics[width=0.13\textwidth]{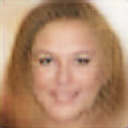}}%
\hspace{9pt} 
\subfigure[Epoch 100]
{\includegraphics[width=0.13\textwidth]{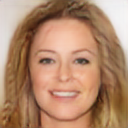}}%
\hspace{9pt}
\subfigure[Epoch 200]
{\includegraphics[width=0.13\textwidth]{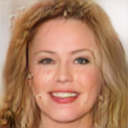}}%
\hspace{9pt}
\subfigure[Epoch 250]
{\includegraphics[width=0.13\textwidth]{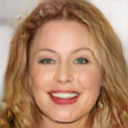}}%
\hspace{30pt}
\subfigure[Target]{\includegraphics[width=0.13\textwidth]{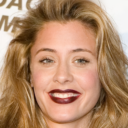}}\\

\subfigure[Epoch 12] {\includegraphics[width=0.13\textwidth]{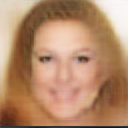}}%
\hspace{9pt}
\subfigure[Epoch 100] {\includegraphics[width=0.13\textwidth]{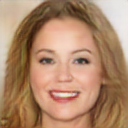}}%
\hspace{9pt}
\subfigure[Epoch 200] {\includegraphics[width=0.13\textwidth]{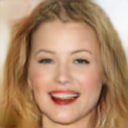}}%
\hspace{9pt}
\subfigure[Epoch 250] {\includegraphics[width=0.13\textwidth]{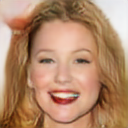}}
\hspace{30pt}
\subfigure[Target]{\includegraphics[width=0.13\textwidth]{female2_target.png}}
\caption{Reconstruction of the target image [\textbf{(e)}, \textbf{(j)}] using IntroVAE with fixed loss weighting from Huang et al. \cite{Huang2018} [images \textbf{(a-d)}] vs our adaptive loss weighting with SoftAdapt \textbf{(f-i)}]} 

\label{fig:female2SA}
\end{figure*}

Tables \ref{tab:female1} and \ref{tab:female2} demonstrate the quantitative comparisons: Peak signal-to-noise ration (PSNR), Structural Similarity Index (SSIM) and Naturalness Image Quality Evaluator (NIQE) for Fig. \ref{fig:female1SA}, \ref{fig:female2SA}. These figures illustrate the IntroVAE reconstruction of random subset of $128 \times 128$ CELEBA dataset using authors' fixed weights versus using our method to find those weights dynamically (Fig. \ref{fig:female1SA}). The training time between the two methods were also very comparable, 1411.489043 minutes for fixed weights vs.  1413.112740 minutes with SoftAdapt. 
\begin{table*}[!ht]
\caption{ Comparison between SoftAdapt and fixed loss weights for Fig. \ref{fig:female1SA} (boldface indicates better performance). } \smallskip
\label{tab:female1}
\begin{tabular}{l|llll|llll|}
\cline{2-9} 
 & \multicolumn{4}{ |c| }{\textbf{Ours (SoftAdapt)}}&  \multicolumn{4}{ |c| }{\textbf{Huang et. al.}}\\

\cline{2-9} 
                           &  \multicolumn{1}{l|}{\textit{Epoch 12}} & \multicolumn{1}{l|}{\textit{Epoch 100}}        & \multicolumn{1}{l|}{\textit{Epoch 200}} & \textit{Epoch 250} & \multicolumn{1}{l|}{\textit{Epoch 12}} & \multicolumn{1}{l|}{\textit{Epoch 100}} & \multicolumn{1}{l|}{\textit{Epoch 200}} & \textit{Epoch 250} \\ \hline
\multicolumn{1}{|l|}{\textit{SSIM}} & \multicolumn{1}{l|}{\textbf{0.7752}}   & \multicolumn{1}{l|}{\textbf{0.8331}}           & \multicolumn{1}{l|}{\textbf{0.8100}}    & \textbf{0.8473}    & \multicolumn{1}{l|}{0.7551}   & \multicolumn{1}{l|}{0.8018}    & \multicolumn{1}{l|}{0.7847}    & 0.7838    \\ \hline
\multicolumn{1}{|l|}{\textit{PSNR}} & \multicolumn{1}{l|}{\textbf{21.5620}}  & \multicolumn{1}{l|}{\textbf{23.3376}}          & \multicolumn{1}{l|}{\textbf{23.8525}}   & \textbf{23.9272}   & \multicolumn{1}{l|}{21.4471}  & \multicolumn{1}{l|}{23.0899}   & \multicolumn{1}{l|}{23.2070}   & 22.2415   \\ \hline
\multicolumn{1}{|l|}{\textit{NIQE}} & \multicolumn{1}{l|}{18.8726}  & \multicolumn{1}{l|}{\textbf{18.8715}}          & \multicolumn{1}{l|}{18.8720}   & \textbf{18.8705}   & \multicolumn{1}{l|}{\textbf{18.8725}}  & \multicolumn{1}{l|}{18.8731}   & \multicolumn{1}{l|}{\textbf{18.8711}}   & 18.8714 \\ \hline
\end{tabular}
\end{table*}
\vspace{0.1cm}

\begin{table*}[!ht]
\caption{ Comparison between SoftAdapt and fixed loss weights for Fig. \ref{fig:female2SA} (boldface indicates better performance). } \smallskip
\label{tab:female2}
\begin{tabular}{l|llll|llll|}
\cline{2-9} 
 & \multicolumn{4}{ |c| }{\textbf{Ours (SoftAdapt)}}&  \multicolumn{4}{ |c| }{\textbf{Huang et. al.}}\\
\cline{2-9} 
                          &  \multicolumn{1}{l|}{\textit{Epoch 12}} & \multicolumn{1}{l|}{\textit{Epoch 100}}        & \multicolumn{1}{l|}{\textit{Epoch 200}} & \textit{Epoch 250} & \multicolumn{1}{l|}{\textit{Epoch 12}} & \multicolumn{1}{l|}{\textit{Epoch 100}} & \multicolumn{1}{l|}{\textit{Epoch 200}} & \textit{Epoch 250} \\ \hline
\multicolumn{1}{|l|}{\textit{SSIM}} & \multicolumn{1}{l|}{0.7940}   & \multicolumn{1}{l|}{\textbf{0.8260}}           & \multicolumn{1}{l|}{\textbf{0.8306}}    & \textbf{0.8303}    & \multicolumn{1}{l|}{\textbf{0.8042}} & \multicolumn{1}{l|}{0.8167}    & \multicolumn{1}{l|}{0.8214}    & 0.8083    \\ \hline
\multicolumn{1}{|l|}{\textit{PSNR}} & \multicolumn{1}{l|}{18.0110}  & \multicolumn{1}{l|}{\textbf{19.4724}}          & \multicolumn{1}{l|}{\textbf{20.0634}}   & \textbf{19.8027}   & \multicolumn{1}{l|}{\textbf{18.7680}}  & \multicolumn{1}{l|}{19.0574}   & \multicolumn{1}{l|}{19.1012}   & 19.3225   \\ \hline
\multicolumn{1}{|l|}{\textit{NIQE}} & \multicolumn{1}{l|}{\textbf{18.8700}}  & \multicolumn{1}{l|}{18.8740}          & \multicolumn{1}{l|}{\textbf{18.8744}}   & \textbf{18.8750}   & \multicolumn{1}{l|}{\textbf{18.8730}}  & \multicolumn{1}{l|}{\textbf{18.8731}}   & \multicolumn{1}{l|}{18.8763}   & 18.8756   \\ \hline
\end{tabular}
\end{table*}

\newpage 
\textbf{  }
\newpage 
\textbf{  }

\subsection{Sparse Autoencoders}

Autoencoders (AEs) are models that aim to reconstruct the input as the output. These networks are comprised of two parts: 1) \textit{Encoder}, a neural network where the data is mapped to a latent space, typically of a smaller dimension than the input. 2) \textit{Decoder}, a neural network where the latent space is mapped back to the original dimension of the network, and, in an optimal case, an exact reconstruction of the input to the encoder. A general autoencoder has the loss of form $$ L(x,\hat{x}) = L(x,g\left(f(x)\right)) $$ where $x$ represents the data, $\hat{x}$ denotes the data reconstruction and $f(x), g(x)$ are the mappings of the encoder and the decoder respectively. To test the performance of SoftAdapt, we trained an AE to reconstruct the MNIST digits \cite{MNIST,MNISTDataBase} with a loss function : $$L(x,\hat{x}) = \frac{1}{n} \sum_{i=1}^n (x_i - \hat{x}_i)^2 + \lambda \sum_{i=1}^m |a_i^{(h)}|,$$ where the added $L_1$ regularization tries to penalize the absolute value of activation layer for a sample $i$ in layer $h$; this is known as a sparse autoencoder \cite{Makhzani2013} since the $L_1$ regularization on the activation of the hidden layers enforces activation of only a few neurons when a sample is inputted. Normally, the hyper-parameter $\lambda$ is tuned to control the effect of the regularization by trial and error. We used SoftAdapt to dynamically adjust the effects of the penalty depending on the performance of each component ($MSE$ and $L_1$ regularization) and the network as a whole. The new loss function using SoftAdapt becomes: 

\begin{equation}
 L(x,\hat{x})^{(k+1)} = \alpha_1^{(k)} \frac{1}{n} \sum_{i=1}^n (x_i - \hat{x}_i)^2 + \alpha_2^{(k)} \sum_{i=1}^m |a_i^{(h)}|
\end{equation}

where $k$ denotes the current iteration and  
\begin{equation} \alpha_i^{(k)} = SoftAdapt\left(\left[\frac{1}{n} \sum_{i=1}^n (x_i - \hat{x}_i)^2 \right]^{(k)}, \left[\sum_{i=1}^m |a_i^{(h)}|\right]^{(k)} \right) \end{equation}

\begin{figure}[H]
\centering
\includegraphics[width=8cm]{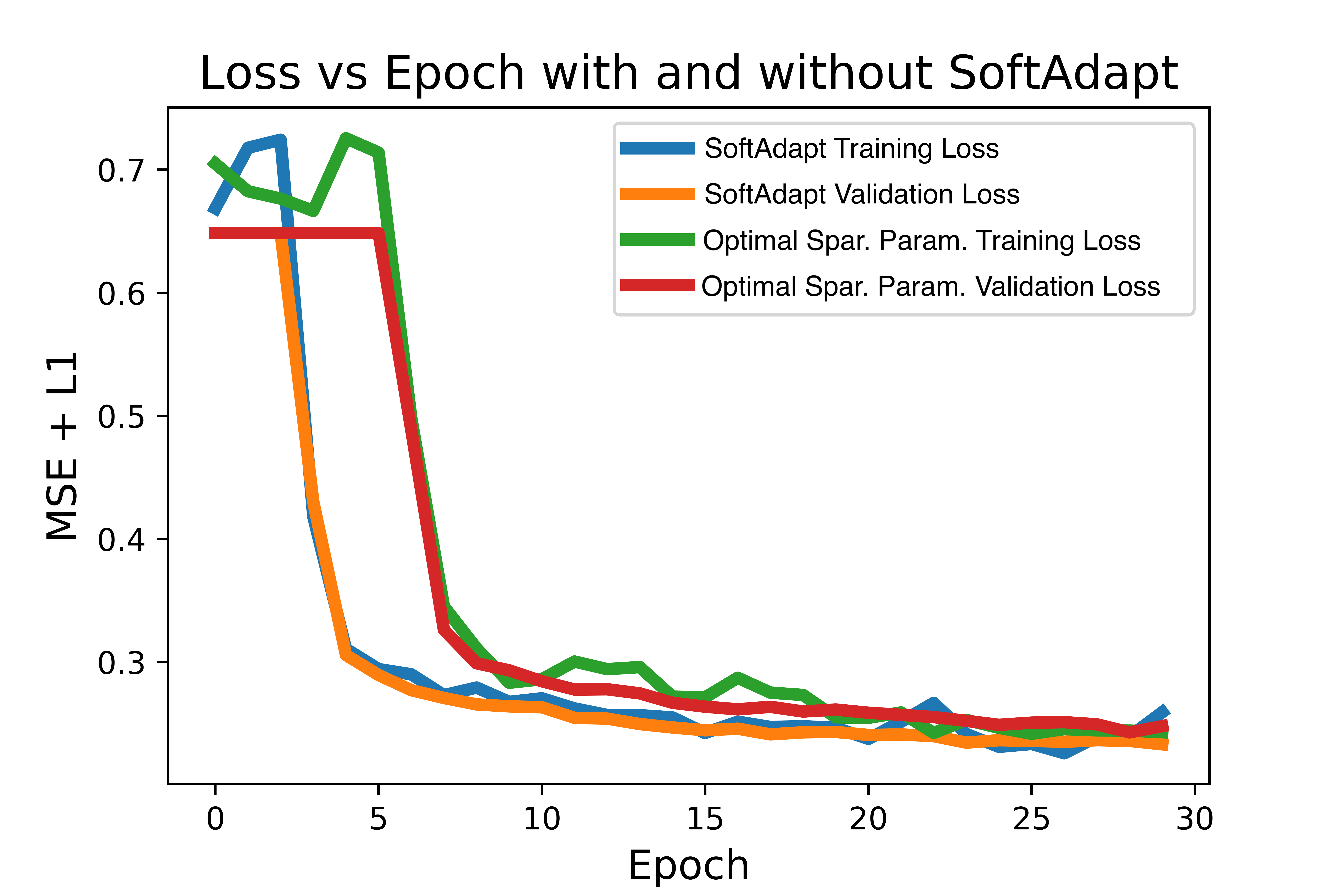}
\bigskip
\caption{Loss vs. epoch for a sparse autoencoder trained with $\lambda = 10^{-4}$ (``optimal")  against using SoftAdapt for weight balancing. Our method performs better throughout training, although the traditional method is comparable to ours for a larger number of epochs.}
\label{fig:AE}
\end{figure}

\begin{figure*}[!ht]
     \centering
     \subfigure[Trained with Fixed $\lambda = 10^{-4}$] {\includegraphics[width=0.3\textwidth]{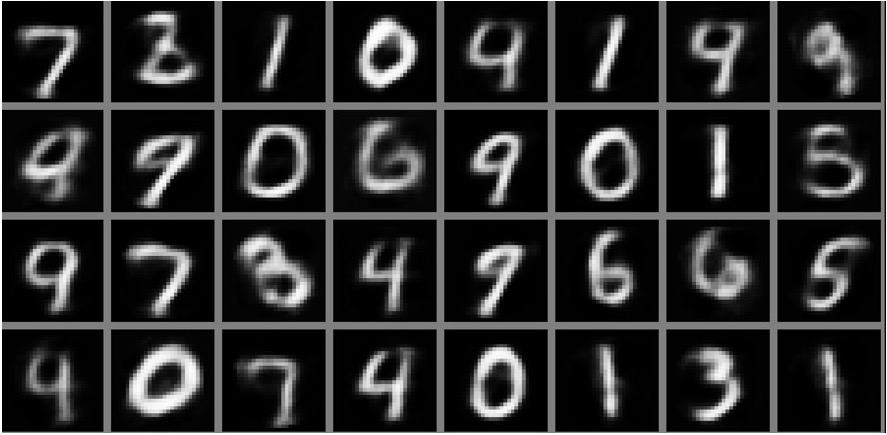}}%
     \hfill
     \subfigure[Trained with SoftAdapt] {\includegraphics[width=0.3\textwidth]{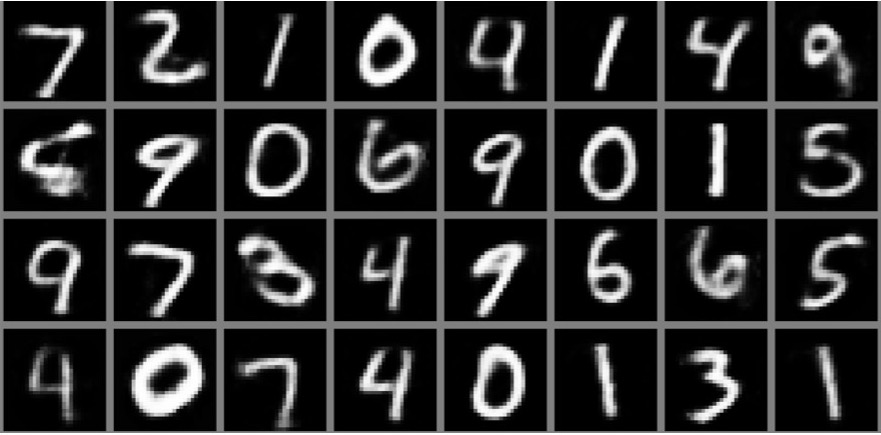}}%
     \hfill
     \subfigure[Target] {\includegraphics[width=0.3\textwidth]{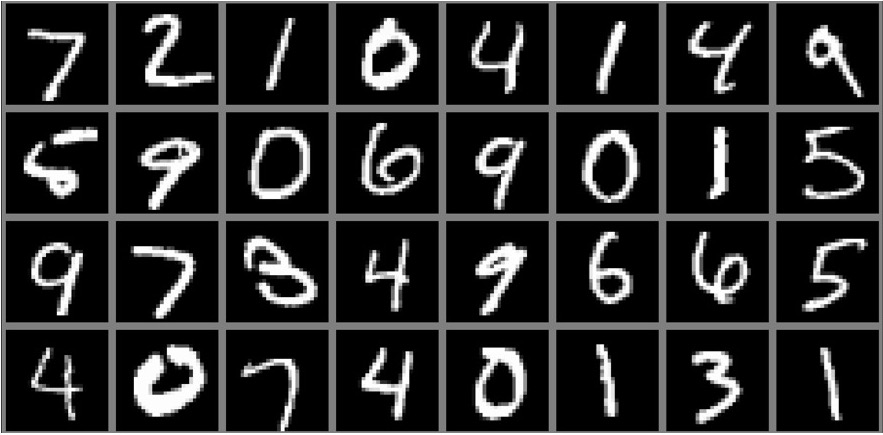}}%
     \caption{Reconstruction of a random set of MNIST \cite{MNISTDataBase} digits from the testing data with a sparse autoencoder using our algorithm \textit{SoftAdapt}. With SoftAdapt, there is no need to find the sparsity parameter $\lambda$ explicitly and by hand. The performance of the autoencoder using our algorithm is comparable to training the network with a fixed optimal value of $\lambda$ found by trial and error.}
     \label{fig:sparse}
 \end{figure*}

Fig. \ref{fig:AE} shows that our method keeps the loss for both training and validation data lower than the traditional ``optimal" $\lambda$, and Table \ref{tab:AE} demonstrates that our reconstructions have a higher classification throughout training than the fixed optimal $\lambda$. We also show that our method is qualitatively comparable to training the network using the optimal $\lambda = 10^{-4}$ from the beginning (Fig. \ref{fig:sparse}). It is worthy to note that the optimal $\lambda$ is found through trial and error, in our case using a grid search which is expensive, but with SoftAdapt no prior knowledge of the values of $\lambda$ is required. Details about network architecture and other hyper-parameters are presented in the \textit{Supplemental Material} section.

 \begin{table*}[t]
\centering
\caption{Classification and time comparison between adaptive weights (ours) and "optimal" $\lambda$ (found manually) in Sparse Autoencoder with loss $L = MSE( \cdot ) + L_1 Regularization $}
\label{tab:AE}
\begin{tabular}{l|llll|llll|}
\cline{2-9} 
 & \multicolumn{4}{ |c| }{\textbf{Ours (SoftAdapt)}}&  \multicolumn{4}{ |c| }{\textbf{Fixed $\lambda = 10^{-4}$}}\\

\cline{2-9} 
                           &  \multicolumn{1}{l|}{\textit{Epoch 2}} & \multicolumn{1}{l|}{\textit{Epoch 5}}        & \multicolumn{1}{l|}{\textit{Epoch 15}} & \textit{Epoch 30} & \multicolumn{1}{l|}{\textit{Epoch 2}} & \multicolumn{1}{l|}{\textit{Epoch 5}} & \multicolumn{1}{l|}{\textit{Epoch 15}} & \textit{Epoch 30} \\ \hline
\multicolumn{1}{|l|}{\textit{PCC}} & \multicolumn{1}{l|}{$11\%$}   & \multicolumn{1}{l|}{\textbf{75 $\%$}}           & \multicolumn{1}{l|}{\textbf{87 $\%$}}    & \textbf{88 $\%$}    & \multicolumn{1}{l|}{$11\%$}   & \multicolumn{1}{l|}{52$\%$}    & \multicolumn{1}{l|}{69$\%$}    & 82$\%$    \\ \hline

 \multicolumn{1}{|l|}{\textit{Time}} & \multicolumn{4}{ |c| }{8.986135 Minutes}&  \multicolumn{4}{ |c| }{\textbf{7.939554 Minutes}}\\ \hline 
 
\end{tabular}
\end{table*}

\section{Conclusion}

We have presented a set of optimization add-ons for weighting the importance of different components adaptively in multi-part objective functions. By adjusting the weights dynamically, the training can become much easier and faster since no prior knowledge of the network is required, \ie no pre-training or grid search is needed. We outlined multiple variants of our Softmax-inspired algorithm and described the suitable application for each one. The first variant of SoftAdapt is a Softmax function where the rate of change is the input, which serves as a performance measure of each part. This is useful when the components of the loss function have the same order of magnitude (\eg various euclidean norms). The second variant uses the magnitude of each part of the loss function as well as the rate of change, which gives the most improvement when the values of the objective functions are on different scales. This variant also has the advantage of assigning smaller weights to the loss functions that are close to their minima, even with large rates of change, and putting more importance on the rest of the objective functions. Our last variant uses normalized rates of change to ensure a better distribution of weights when the slopes possess vastly different scales. It is important to note that the second and third variants may be used together if needed. Our SoftAdapt algorithm is implemented in one easy-to-use package available online on the authors' websites (not included due to the blind review). Our results show that our algorithm works well in practice for a wide spectrum of problems in machine learning, such as image reconstruction and synthetic data generation, as well as general gradient decent optimizations where scaling is an issue.


\bigskip
\bigskip
\bigskip
\bigskip


\section*{Acknowledgments}       
 
 We would like to acknowledge Omar DeGuchy, Radoslav Vuchkov and Alina Gataullina for their constructive comments regarding our methods and writings. We would like to thank Fred Garber, Olga Mendoza-Shrock, Jamison Moody, Oliver Nina, Alexis Ronnebaum and Suzanne Sindi for their feedback and support. We also appreciate the computation resources provided by the University of California, Pacific Research Platform and the Wright State University to the authors in conducting this research.
 
\bigskip
\bigskip
\bigskip
\bigskip
\bigskip
\bigskip
\bigskip
\bigskip
\bigskip
\bigskip
\bigskip
\bigskip
\bigskip
\bigskip
\bigskip
\bigskip
\bigskip
\bigskip
\bigskip
\bigskip
\bigskip
\bigskip

\newpage
\newpage

\bibliography{new} 
\nocite{*}

\bibliographystyle{ieee_fullname} 

\end{document}